\documentclass[conference]{IEEEtran}
\ifCLASSINFOpdf
\else
\fi
\usepackage{multirow}
\usepackage{tabu}
\usepackage{graphicx}
\usepackage{multirow}
\usepackage{array}
\usepackage{amsmath}

\begin{document}
%
\title{A Comparative Study of the Clinical use of Motion Analysis from Kinect Skeleton Data}


\author{\IEEEauthorblockN{Sean Maudsley-Barton\IEEEauthorrefmark{1},
		Jamie McPhee\IEEEauthorrefmark{2},
		Anthony Bukowski\IEEEauthorrefmark{1}, 
		Daniel Leightley\IEEEauthorrefmark{3} and 
		Moi Hoon Yap\IEEEauthorrefmark{1}}
	\IEEEauthorblockA{\IEEEauthorrefmark{1}School of Computing, Mathematics and Digital Technology\\
		Manchester Metropolitan University, Manchester, M1 5GD, UK\\
		Email: sean.maudsley-barton@mmu.ac.uk§}
	\IEEEauthorblockA{\IEEEauthorrefmark{2}School of Heathcare Science\\
		Manchester Metropolitan University, Manchester, M1 5GD, UK}
	\IEEEauthorblockA{\IEEEauthorrefmark{3}King's Centre for Military Health Research \\
		King's College, London, WC2R 2LS, UK}}


%


\maketitle

\begin{abstract}
The analysis of human motion as a clinical tool can bring many benefits such as the early detection of disease and the monitoring  of recovery, so in turn helping people to lead independent lives. However, it is currently under used. Developments in depth cameras, such as Kinect, have opened up the use of motion analysis in settings such as GP surgeries, care homes and private homes. To provide an insight into the use of Kinect in the healthcare domain, we present a review of the current state of the art. We then propose a method that can represent human motions from time-series data of arbitrary length, as a single vector.  Finally, we demonstrate the utility of this method by extracting a set of clinically significant features and using them to detect the age related changes in the motions of a set of 54 individuals, with a high degree of certainty (F1-score between 0.9 - 1.0).  Indicating its potential application in the detection of a range of age-related motion impairments.
\end{abstract}


%
\IEEEpeerreviewmaketitle

\section{Introduction}
The analysis of human motion from video has a long history, beginning in the 1970s with the experiments of Johansson \cite{Johansson1973}, which demonstrated people can recognise a wide range of human motions, even if only the joints of an actor are highlighted.  This work led directly to the large multi camera, marker based, motion capture systems, still regarded by many as the gold standard for motion capture (mocap).  The introduction of the Kinect depth camera, in 2011, heralded a new age of inexpensive markerless motion capture. Most importantly, thanks the work of Shotton et al. \cite{Shotton2011}, Kinect reduces the depth image to a stick skeleton, who's joints while not strictly anatomically correct have been demonstrated to have a high correlation with its marker-based kinsman \cite{galna2014accuracy}.  Originally conceived to rival the Wii, Kinect has now found applications in areas as diverse as facial analysis \cite{Kendrick2017}, surveillance, security and health.

In the nearly 6 years since its introduction, much effort has been put into using the Kinect to analyse human motion.  Often but not always allied to some form of automatic classification and/or quantification.  Initially this work used classical machine learning techniques such as Random Forests and Support Vector Machine (SVM).  Although these techniques have provided excellent results \cite{Leightley2016},  the newer deep learning techniques offer the possibility of detecting finer detail.  This paper provides a comparison of traditional machine learning approaches and the newer deep learning techniques. their relative success so far and details our own investigations using the K3Da dataset.  Our initial findings demonstrate that deep learning can outperform traditional methods.

\subsection*{Scope of review}
Human motion analysis is a broad topic, it encompasses areas of facial movement, hand movement and sign language recognition, rehabilitation and many others. This study limits itself to the use of skeleton data for the analysis of motions in a clinical setting.  Motion analysis is used by health professionals in the assessment of a range of diseases and conditions which affects both young and old.  Prior to the introduction of Kinect, the cost and availability of marker-based systems meant that only a small section of the population had access to this type of assessment.  The majority of assessments were, and still are done subjectively, by direct observation. Or rely on patients filling in a questionnaire.  Even after the introduction of Kinect, many studies \cite{Gianaria2016KinectGait}, \cite{Stone2012} and applications still require an expert to interpret the results.  This is where machine intelligence can play a role, by providing quick and objective analysis of human movement.

\section{Related Work}
From the introduction of Kinect in 2011, the possibilities for use in healthcare were recognised.  Some researchers have concentrated on rehabilitation while others work on the quantification of movement.  Before Kinect, attempts had been made to use RGB cameras in rehabilitation schemes.  These systems used colour segmentation to identify limbs.  This made them very sensitive to light, shadows and distance from the camera.  Kinect's technology is indifferent to these issues. Many studies have used Kinect in rehabilitation, sometimes referred to as the gamification of physiotherapy.  It both reduces the strain on therapeutic resources and encourages individuals, to repeat their beneficial exercises \cite{Chang2011}, \cite{Luna-Oliva2013}, \cite{Ortiz-Gutierrez2013a}.

Stone et al. \cite{Stone2011} used Kinect in gait analysis.  They used it to calculate gait velocity and length in the lab.  Followed up, a year later, by a proof that Kinect cameras can be used to continuously monitor gait in the home \cite{Stone2012}. Dolatabadi et al. \cite{Dolatabadi2013} wrote a case study, detailing how they used features derived from Kinect to unobtrusively monitor recovery from hip replacements, in the home.   Pu et al. \cite{Pu2015} used Kinect to assess balance in a selection of 100 people.  These studies calculated hand crafted features from the Kinect data and analysed them to demonstrate their significance.
 
As the popularity of Kinect has grown a cottage industry in validating Kinect against the gold standard for mocap has sprung up.  One of the most extensive comparisons was carried out by  a large team, led by Otte \cite{Otte2016}.  They validated both V1 and V2 of Kinect and found there was high agreement between both versions, however there was some spread between different joints, the head joint being the most accurately tracked and feet joints being the least. Yang et al. \cite{Yang2014} demonstrated a small, linear discrepancy between the Centre of Mass (CoM), calculated using Kinect and the ground truth.  As this discrepancy is linear, it can easily be accommodated or just ignored for calculations which look at the change of a quantity over time, for example postural sway.

The studies,  highlighted above, rely on expert analysis by humans.  There has been no attempt to automate the process or leverage machine intelligence.  The next set of papers look at, do just that.

\subsection{Machine Intelligence}
The most basic application of machine intelligence is to apply a heuristic approach, as  Bigy et al. \cite{Bigy2015} did in their real-time system which can detect Freeze of Gait (FoG) a problematic symptom of Parkinson's disease as well as falls and tremors \cite{Bigy2015}.  Their simple rules-based approach makes this easily applied in real time.

Alexiadis et al. \cite{Alexiadis2011}, took a more sophisticated approach.  in their work related to an automatic method for evaluating a dancer's performance.  They developed three metrics and techniques to temporally align the movements of amateur dancers with ground truth provided by professionals. Inspired by this work, Su  et al. \cite{Su2014} developed a system which uses Dynamic Time Warping (DTW), a method previously used in relation to hand writing and audio recognition, to compare the movements of patients in their everyday life to a standard set of movements.  The aligned sequences are then classified using an Adaptive Neuro-Fuzzy Inference System (ANFIS) which uses a combination of artificial neural network and fuzzy logic.

A drawback with DTW is that it does not perform well for periodic movements like waving.  Wang et al. \cite{Wang2012} proposed the pairwise encoding of the relative joints, producing a much more discriminating feature set.  This process relies on the normalisation of skeleton graphs so that all joint offsets are relative to those in the first frame.  This type of normalisation has become the de-facto pre-processing step when calculating features from Kinect skeletons.

Gabel et al. \cite{Gabel2012} extracted a feature set which expanded on gait analysis to include arm kinematics and additional body features namely, Centre of Mass (COM), Direction of Progress (DoP) and Acceleration.  These measures have become common features of many skeleton based studies.  A combination of Multiple Additive Regression Trees (MART) and a state model were used to predict the gait cycle.  However, machine learning was not used to produce a predictive model for gait quality. 

Greene et al. \cite{Greene2010} had demonstrated the use of logistic regression to quantify falls risk, with data derived from kinematic sensors but the groups using Kinect had yet to venture in to prediction.

Cary et al. \cite{Cary2014} decomposed each skeleton generated by Kinect into a 17-value, feature vector of spherical coordinates.  This was calculated, in relation to torso bias, a 3-value vector produced by applying Principal Component Analysis (PCA) to the torso joints. They used these features to train an Artificial Neural Network (ANN) to recognise a range of movements.

Kargar et al. \cite{Kargar2015}, extracted both gait and angle-based anatomical features from skeletons in order to quantify a person's physical mobility.  To achieve this they used a SVM.

In 2016, Leightley et al. \cite{Leightley2016} detailed the first end-to-end pipeline for the automated analysis and quantification of human movement, using Kinect.  The system uses a variety of motion analysis techniques to first extract clinically significant features and then quantify them using SVM.

\subsection{Deep Learning}
In the last few years, there have been a flurry of papers that use deep learning to either extract features or analyse human movement.  LeCun et al. \cite{LeCun2004} demonstrated the ability of Convolutional Neural Networks (CNN) to outperform SVM models for static images. 

The first 3D CNN was created by Ji et al. \cite{ji2013} to address the issue of automatic feature extraction and recognition from RGB airport footage.  Ijjina et al. \cite{Ijjina2014} demonstrated the use of 3D CNNs with motion capture, Leightley et al. \cite{Leightley2017} applied a similar network structure to clinically significant motions, captured by Kinect.  They demonstrated significant improvement over the previous best method (SVM) in differentiating between good and unstable motions. 

In this paper we demonstrate a hybrid approach to deep learning.  We use manually extracted clinically significant features and then use a deep, fully connected neural network to classify the motion.  When comparing this approach to random forests and SVM.  The deep neural network out performs the other classifiers.

\subsection{Datasets} 
Datasets are a constant issue when training deep neural networks.   The largest dataset currently available for RGB data is the Sports-1M dataset \cite{Google2015}, which contains one million sports videos culled from YouTube and separated into 487 classes \cite{Jardim2017}.  RGB-D (depth) datasets are much smaller.  The largest currently available being NTU RGB+D \cite{Shahroudy2016} with 56,880 recordings of 50 individuals, carrying out 60 classes of movement.  Zhang et al. \cite{Zhang2016}, identified a total of 44 RGB-D datasets.  7 of these concentrate on tracking multiple individuals and 10 use an array of multiple cameras.  Of the remaining 27, only 2 are captured in laboratory conditions.  However, the movements they capture are not clinically relevant.

To address the issue of small datasets, the creation of synthetic data has become popular, indeed the random forest used to generate Kinect skeletons uses synthetic data in its training set.  Validation issues have made researchers in clinical studies, shy away from this approach to bulk out small datasets.  Zhang et al. \cite{Zhang2017} proposed a method for synthesising data, within the network that improved the recognition rate of motions.  It is reported that this form of enhancement, overcomes the limitations of the human defined pre-processing approaches. In time  this type of approach may offer answers to those objections found in the clinical sphere.  To this end we demonstrate an automatic method for the production of a family of vectors that can fully describe a motion of any length from a time-series of Kinect skeletons.

Currently, the K3Da dataset, highlighted in \cite{Firman2016} is the only dataset that contains RGB-D data of clinically important actions.  It contains 576 recordings of 54 individuals completing the 13 movements from the Short Physical Performance Battery (SPPB) \cite{guralnik1994SPPB}.  This is the dataset used in our investigations.

\section*{Methods}
Someone with good postural control, when standing, will be able to keep their Centre Of Mass (CoM) over their Base of Support (BoS).  Someone with poor postural control, their CoM will more often be outside their BoS and they will initiate more frequent and larger postural corrections, evident as higher postural sway, or poor balance.  The corrective actions are initiated through motor control pathways achieved via joint moments applied around the ankle, knee and the hip \cite{Torres-Oviedo2010}.   The extent to which these different joints are active depends on the extent of the postural challenge.  When the CoM during quiet standing irrecoverably deviates from the base of support, the person will take a step to rescue from falling.  During two-legged quiet standing, the base of support is stable in the medio-lateral (ML) directions, so the ankle and hip strategies \cite{Horak1989} mainly work to minimise  instability in the anterior-posterior (AP) axis.  In less stable foot positions, such as one-foot stand, instability around the ankle mainly occurs in the ML-axis. To measure deviations in the AP axis, we calculate the body lean angle, that is the Euler angle between the ground plane and the middle of the spine.  Norris et al. \cite{Norris2005} point to the loss of postural control also resulting in movements in the ML axis.  To account for this, we record the position of the spine joints in the ML axis, ignoring movements in the AP axis to reduce noise.  Together these features give us a measure of postural sway.  

In addition to measuring sway,  we directly estimate the CoM position, Leightley et al. \cite{Leightley2016-Challenging} found this to be a useful measure of steadiness.  We complete our features by calculating Euclidean distance between the base of the spine and the head and the Euler angle between the base of the spine and the neck. Ejupi et al. \cite{Ejupi2015} discusses the importance of visual and somatosensory systems, as well as the vestibular system in maintaining balance.  To factor these items into our study, we chose tasks that would challenge all three systems.  The movements used and the reason for their inclusion are detailed in table \ref{tab:movments}.
\begin{table}[ht!]
	\centering
	\caption{Movements used in this work, along with the rational for inclusion}
	\label{tab:movments}
	\begin{tabular}{|>{\centering\arraybackslash}p{1.6cm}|>{\raggedright\arraybackslash}p{3.1cm}|>{\raggedright\arraybackslash}p{2.7cm}|}
		\hline
		\textbf{Movement Name}& {\textbf{Description}}&                                    
		{\textbf{Reason for inclusion}}                                                                                                                                                                              \\ \hline
		\textbf{Chair Rise}                    &  Starting from a seated position, rise up with legs fully extended, then sit down again, arms are held across the chest. Repeat five times, as quickly as possible. & This measure is indicative of leg muscle power, associated with falls and physical impairment \cite{Tiedemann2008}.                                                                                                                        \\ \hline
		\textbf{Stand, 2 Feet(eyes open)}       & Standing feet close together, eyes open and arms extended parallel to the floor. Test is terminated after 10 seconds.                                                                                                                        & This is used to provide an indication of postural sway in a quiet stance, two feet on the floor.                                                                                                                                \\ \hline
		\textbf{Stand, 2 Feet(eyes closed)}     & Standing feet close together, eyes closed and arms extended parallel to the floor. Test is terminated after 10 seconds.                                                                                                                        & Removing the visual cues make the participant rely on vestibular and somatosensory feedback.  Both these systems are effected by age \cite{Ejupi2015}. \\ \hline
		\textbf{1 leg balance (open eyes)}   & Standing on one leg, 6 inches off the ground, arms extended horizontally, eyes open.  Test terminated after 10 seconds or when the second leg touches the ground.                                                                                                 & Balancing on one leg provides a small base of support, inducing postural sway.                                                                                                                                     \\ \hline
		\textbf{1 leg balance (closed eyes)} & Standing on one leg, 6 inches off the ground, arms extended horizontally, eyes closed.  Test terminated after 10 seconds or when the second leg touches the ground.                                                                   & During this exercise, the individual must use the vestibular and somatosensory systems to maintain balance.                                                 \\ \hline
	\end{tabular}
\end{table}

In this work we used the skeleton data from the K3Da dataset \cite{Leightley2015K3Da}, which is the largest data set of it's kind.  It consists of, 26 young and middle aged people (18-48 years, 17 male and 9 female) and 28 older age people (61-81 years, 14 male and 14 female) carrying out the SPPB.  None of the participants have any non-age-related movement issues.

\begin{figure*}[!ht]
	\centering
	\includegraphics[width=0.53\linewidth]{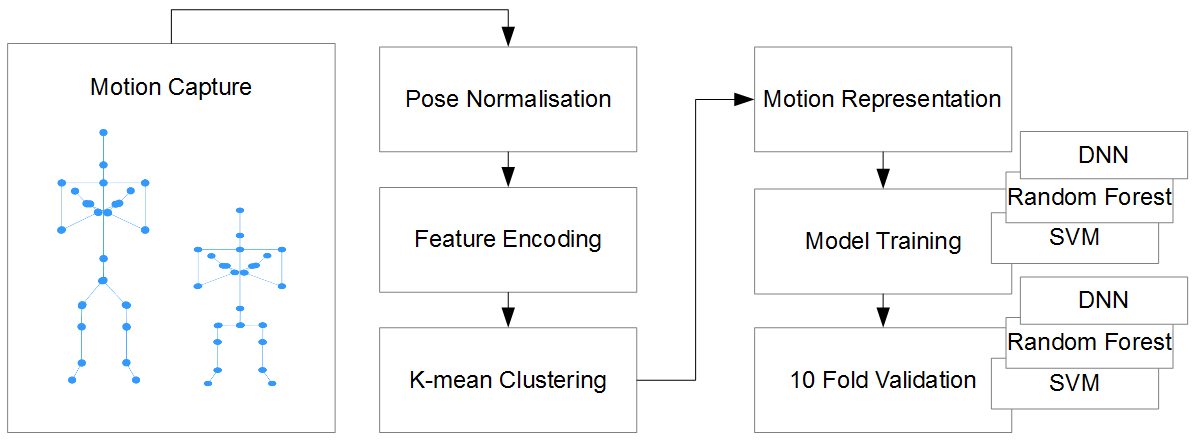}
	\caption{Pipeline to extract medically significant features from a RGB-D video stream}
	\label{fig:pipeline}
\end{figure*}

Using the methods outlined in fig. \ref{fig:pipeline}, we automated the process of feature encoding and classification of an individual, based upon their motion alone.  The data processing step was carried out using Matlab.  Matlab was also used for classification by traditional machine learning.  The Theano framework was used to build the neural network.  The following section covers each step in detail.

\subsection*{Pose Normalisation}
From the skeleton data, a series of matrices were constructed, one for each frame of the movement (30 fps).  The skeletons were normalised by aligning all frames to the Spine Base joint of the first frame, using equation \ref{eq:erl}.

\begin{equation} \label{eq:erl} 
p_{n,i}(x,y,z)^*= P_{n,i}(x,y,z) - P_{spinebase},1(x,y,z)
\end{equation}

\subsection*{Feature Encoding}
After normalisation, a set of features, shown in 
Table \ref{tab:features}, was calculated.   Although the Kinect camera provides coordinates for 25 joints, we found that features derived from just the torso joints, as defined in \cite{Leightley2016}, plus a calculation of CoM were enough to discriminate the differences in motions of young and older people.


\begin{table}[!ht]
	\caption{Features extracted}
	\label{tab:features}
	\begin{tabular}{|l|>{\arraybackslash}p{2.5cm}|l|}
		\hline
		\textbf{Feature}          & \textbf{Description}                                                    & \textbf{Vector length} \\ \hline
		Euclidean Distance        & Distance between Spine base and Head                                    & 1                      \\ \hline
		Euler angle               & Angle between Spine base and Neck                                       & 1                      \\ \hline
		Body Lean Angle (AP axis) & Euler angle between Spine Mid and the ground plain                      & 1                      \\ \hline
		CoM                       & Mean position between Spine mid, Hip left and Hip right                 & 3                      \\ \hline
		Torso ML Axis                   & XY position of the torso joints & 10                     \\ \hline
	\end{tabular}
\end{table}

\textbf{Euclidean Distance}
was calculated between the spine base and head, using equation \ref{eq:ecd}.

\begin{equation} \label{eq:ecd} 
distance= \sqrt{(x_1-x_2 )^2 + (y_1-y_2 )^2 + (z_1-z_2 )^2} 
\end{equation}

\textbf{Euler Angle}
was calculated between the Spine base and Neck joints, using equation \ref{eq:eua}.

\begin{equation} \label{eq:eua} 
\Theta = \arctan{\left( \frac{S.Q}{\left \|S\right\|\left \|Q\right\|} \right)}
\end{equation}

\textbf{Body Lean Angle}
is defined as the Euler Angle between the ground plane and the Spine mid joint.  This represents changes in the AP axis.

\textbf{Centre of Mass (CoM)} of any body, is the mean point that the mass of that body acts.  For a human body standing erect, the centre of mass is located around the navel. CoM was calculated using equation \ref{eq:com}, where J1 = Spine mid, J2 = Hip left, J3 = Hip right, where

\begin{equation} \label{eq:com}
\begin{split}
CoM_x=\frac{J1_x+ J2_x  +J3_x}{3} \\
CoM_y=\frac{J1_y+ J2_y  +J3_y}{3} \\
CoM_y=\frac{J1_z+ J2_z  +J3_z}{3} \\
\end{split}
\end{equation}

\textbf{ML Axis} This represents the frame-by-frame position of the torso joints in the ML axis.

\subsection*{Motion Representation}

\textbf{K-mean Clustering} is used to convert a time-series of differing lengths (depending on individual recordings), into a set of representative poses of a known length.   The \textit{k} was determined empirically for each type of motion, 5 for chair rise and 2 for all other movements.


The centroid poses of each cluster was identified and extracted.  Next, the centroids were concatenated together in time-order to produce a 1D vector that provides an example of the whole motion.  A label is then added, 1 for young and 0 for older.

Simply choosing the centroid would not provide enough examples to train the models.  More examples were collected, using the following method:-  
\begin{enumerate}
	\item Each member of a cluster were ranked using Euclidean distance  from the centroid.  
	\item The closest 50\% were selected.  To ensure that each vector represents the whole motion, only $n$ number of motion were produced, from each time-series, where $n$ is the number of members in the smallest cluster. 
\end{enumerate}
This method produces a family of feature-sets which are representative of a person's motions.  By providing a family of similar examples, we increase the number of examples retrieved from a time-series, many times, without the  need to resort to synthesising data.  This approach makes the models more robust as they are trained on a diverse but representative  feature-set.

\subsection*{Evaluation}
We compared 3 methods for classification, SVM, Random Forests and Deep Neural Networks.  SVM being widely regarded as the best choice in traditional machine learning for a binary classification, but requires extensive tuning of hyper parameters to achieve top results.  Random Forests, on the other hand are much simpler to train and provide excellent results for both binary and multi-class scenarios.  We chose these methods in our study to allow us to compare the best of traditional machine learning with deep learning.  The neural network consisted of several fully connected layers which learned to separate young from older based on hand crafted features neural networks were not used to extract features.  Feature extraction by neural networks, is an area which will be explored in future work.

\subsection*{Validation}
10-fold cross validation was used to assess the effectiveness of each approach. In K-fold validation every data point gets put into the test set exactly once, and into training set k-1 times. This allows the results to be averaged over the whole dataset.

\section*{Results}
The results, summarised in Table \ref{tab:results}, are similar to those found in the literature for traditional machine learning.  In addition we were able to demonstrate that deep learning is able to outperform traditional methods.

Using our method, that automatically identifies the essence of a motion and then collects many examples of that motion from a time-series of arbitrary length, we were generate enough examples to allow both traditional and deep learning methods to discriminate between young and older people, with a high degree of certainty.  This is demonstrated by high F1-score and Matthews Correlation Coefficient (MCC) scores for all movements, with the chair rise producing the best overall classification.

These results are encouraging. However, we do accept that although, the K3Da is one of the larger depth datasets and the only one currently that contains clinically significant movements, it is still a small dataset when compared with those that contain RGB information.  Consequently, our results may suffer from overfitting.  To address this issue, our future work will involve building a large dataset of clinically significant depth data.

\begin{table}[h]
	\centering
	\caption{All three machine learning algorithms separated the extracted features easily}
	\label{tab:results}
	\begin{tabular}{|>{\centering\arraybackslash}p{1.6cm}|>{\centering\arraybackslash}p{1.2cm}|>{\centering\arraybackslash}p{0.5cm}|>{\centering\arraybackslash}p{0.5cm}|>{\centering\arraybackslash}p{0.6cm}|>{\centering\arraybackslash}p{0.6cm}|>{\centering\arraybackslash}p{0.5cm}|}
		\hline
		\textbf{Action}                                                                                 & \textbf{Model} & \textbf{Acc} & \textbf{Prec} & \textbf{Recall} & \textbf{F1-score} & \textbf{MCC} \\ \hline
		\multirow{3}{*}{\textbf{\begin{tabular}[c]{@{}c@{}}Balance 1 leg, \\ eyes open\end{tabular}}}   & SVM            & 0.999             & 1.000              & 0.998           & 0.999               & 0.998        \\ \cline{2-7} 
		& Random Forest  & 0.998             & 0.998              & 0.998           & 0.998               & 0.998        \\ \cline{2-7} 
		& Deep Learning  & 0.977             & 0.977              & 0.979           & 0.978               & 0.976        \\ \hline
		\multirow{3}{*}{\textbf{\begin{tabular}[c]{@{}c@{}}Balance 1 leg, \\ eyes closed\end{tabular}}} & SVM            & 0.998             & 1.000              & 0.996           & 0.998               & 0.995        \\ \cline{2-7} 
		& Random Forest  & 0.999             & 0.998              & 1.000           & 0.999               & 0.999        \\ \cline{2-7} 
		& Deep Learning  & 0.985             & 0.983              & 0.993           & 0.988               & 0.986        \\ \hline
		\multirow{3}{*}{\textbf{\begin{tabular}[c]{@{}c@{}}Stand 2 feet,\\  eyes open\end{tabular}}}    & SVM            & 0.999             & 1.000              & 0.998           & 0.999               & 0.997        \\ \cline{2-7} 
		& Random Forests  & 0.999             & 1.000              & 0.998           & 0.999               & 0.997        \\ \cline{2-7} 
		& Deep Learning  & 0.991             & 0.994              & 0.994           & 0.994               & 0.985        \\ \hline
		\multirow{3}{*}{\textbf{\begin{tabular}[c]{@{}c@{}}Stand 2 feet,\\ eyes closed\end{tabular}}}   & SVM            & 0.995             & 0.994              & 0.994           & 0.996               & 0.992        \\ \cline{2-7} 
		& Random Forests  & 0.999             & 0.998              & 1.000           & 0.999               & 1.000        \\ \cline{2-7} 
		& Deep Learning  & 0.998             & 0.998              & 0.998           & 0.998               & 0.988        \\ \hline
		\multirow{3}{*}{\textbf{Chair Rise}}                                                            & SVM            & \textbf{1.000}            &\textbf{1.000}             & \textbf{1.000}          & \textbf{1.000}               & \textbf{1.000}        \\ \cline{2-7} 
		& Random Forests  & \textbf{1.000}             & \textbf{1.000}              & \textbf{1.000}           & \textbf{1.000}               & \textbf{1.000}       \\ \cline{2-7} 
		& Deep Learning  & \textbf{1.000}             & \textbf{1.000}              & \textbf{1.000}          & \textbf{1.000}              & \textbf{1.000}        \\ \hline
	\end{tabular}
\end{table}

\section*{Discussion}
There is  a pressing need to develop a portable system that can help in the assessment of physical impairment  and frailty. Currently the assessment of individuals requires a high degree of training and experience, which can lead to inconsistency from one location to the next.

We have taken the first steps in developing a tool which could be used by clinicians in detecting the changes in motion that advance with age.  We demonstrate its utility by separating a random sample of young and older individuals from the K3Da dataset.

\section*{Conclusion and Future Work}     
Our current feature set is working well for a simple two class solution.  However, our future work will concentrate on the prediction of mobility issues.  The current method may lack the power needed to discern the small changes needed to predict future issues.  Hence, we intend to consider the use of deep learning auto-encoders and convolutions, to automatically extract feature sets from motions encoded in depth data.  Using this line of research we may be able to produce a complete pipeline exclusively using deep learning.

We recognise that in order to build a robust end-to-end deep learning solution we need many more examples that exists in the K3Da dataset.  To this end we have taken the first steps towards building a larger dataset of clinically significant motions.  We hope that in turn, this dataset maybe a useful resource for other researchers.

Finally, to fulfil our ambition of developing a tool, useful to clinicians, we must  have a system that works in real time, taking the data feed directly from the Kinect camera.

\bibliographystyle{IEEEtran}
\bibliography{library}

\footnotetext[1]{© 2017 IEEE. Personal use of this material is permitted. Permission from IEEE must be obtained for all other uses, in any current or future media, including reprinting/republishing this material for advertising or promotional purposes, creating new collective works, for resale or redistribution to servers or lists, or reuse of any copyrighted component of this work in other works.}

\end{document}